\pdfoutput=1

\documentclass[11pt]{article}
\usepackage[preprint]{acl}
\usepackage{times}
\usepackage{latexsym}
\usepackage[T1]{fontenc}
\usepackage[utf8]{inputenc}
\usepackage{microtype}
\usepackage{inconsolata}
\usepackage{graphicx}
\usepackage{multirow} 
\usepackage{booktabs}
\usepackage{amsmath}
\usepackage{tikz}
\usepackage{xcolor}
\usepackage{color}
\usepackage{tcolorbox}
\usepackage{colortbl,array}
\usepackage{enumitem}
\usetikzlibrary{tikzmark}
\makeatletter
\newcommand*\myfontsize{%
  \@setfontsize\myfontsize{7}{8}%
}
\makeatother
\newcommand{\mytextbox}[2]{\tikzmarknode[draw=#1,thick,inner sep=2pt]{test}{\myfontsize #2}}
\definecolor{myred}{rgb}{0.7, 0.3, 0.0}
\definecolor{myblue}{HTML}{054488}
\definecolor{mygreen}{HTML}{056b34}
\definecolor{tangerine}{rgb}{0.95, 0.52, 0.0}
\definecolor{amethyst}{rgb}{0.6, 0.4, 0.8}
\definecolor{lasallegreen}{rgb}{0.03, 0.47, 0.19}

\newcommand{\green}[1]{\mytextbox{lasallegreen}{\textbf{\textcolor{lasallegreen}{#1}}}}

\title{GLARE: Agentic Reasoning for Legal Judgment Prediction}
\author{Xinyu Yang$^{1}$,Chenlong Deng$^{1}$, Zhicheng Dou$^{1}$\thanks{Corresponding author.}\\ 
    $^1$Gaoling School of Artificial Intelligence, Renmin University of China \\ 
    \texttt{\{yxygsai,dou\}@ruc.edu.cn} \\
}

\graphicspath{../figures/}
\begin{document}
\maketitle
\begin{abstract}

Legal judgment prediction (LJP) has become increasingly important in the legal field. In this paper, we identify that existing large language models (LLMs) have significant problems of insufficient reasoning due to a lack of legal knowledge. Therefore, we introduce GLARE, an agentic legal reasoning framework that dynamically acquires key legal knowledge by invoking different modules, thereby improving the breadth and depth of reasoning. Experiments conducted on the real-world dataset verify the effectiveness of our method. Furthermore, the reasoning chain generated during the analysis process can increase interpretability and provide the possibility for practical applications. 

\end{abstract}

\section{Introduction}

\begin{figure*}[t]
  \includegraphics[width=\linewidth]{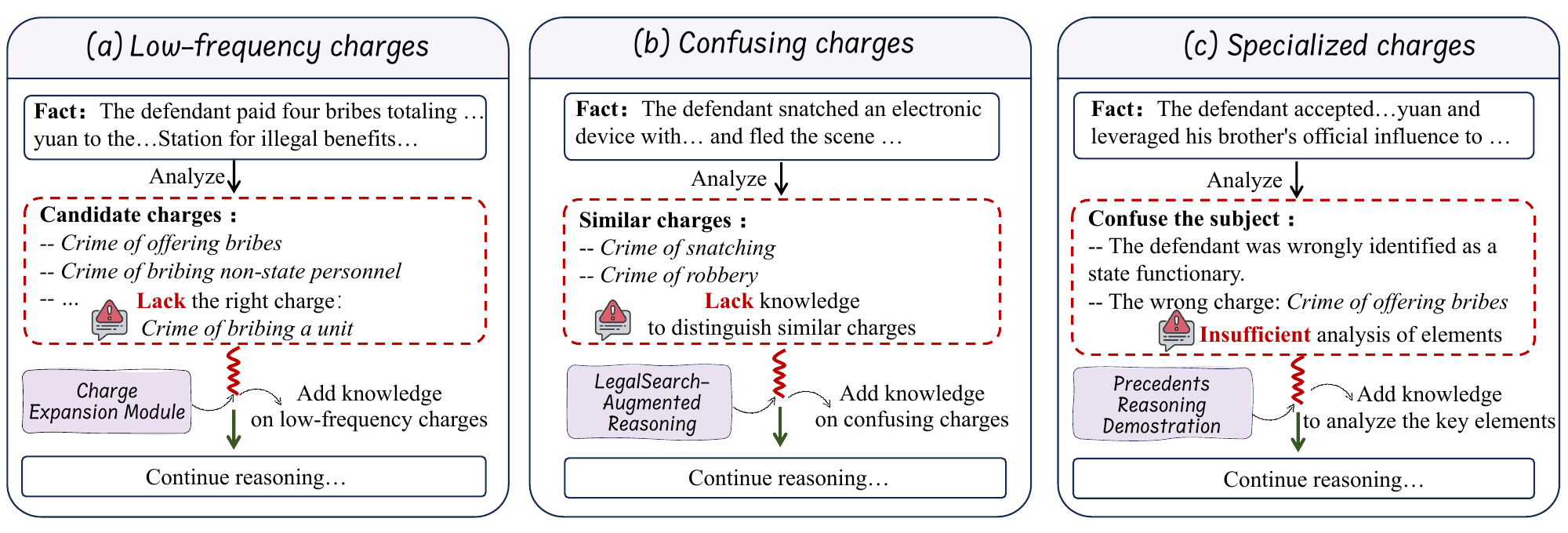}
  \caption{Lack of knowledge in three aspects: (a)\ Lack knowledge of low-frequency charges. (b)\ Lack knowledge of confusing charges. (c)\ Lack knowledge to analyze the key elements of the charges with strong professionalism.}
  \label{fig:intro}
\end{figure*}

Legal judgment prediction (LJP) is an important task in legal natural language processing (NLP), aims to make correct judgment predictions based on the case's fact description~\cite{10.1145/3539618.3591731}. The judgment predictions include law articles, charges, and terms of penalty~\cite{xu2024distinguishconfusionlegaljudgment}. This task not only provides judgment references to lawyers and judges, as well as providing legal consulting services to the general public~\cite{luo2017learning,shulayeva2017recognizing,mcginnis2013great}. 

Recently, large reasoning models (LRMs) have made remarkable progress across in reasoning-intensive tasks, including multi-hop question answering and strategic planning~\cite{wang2024exploring,choi2025thinkclearlyimprovingreasoning}. These models can perform multi-step reasoning that mimics human thinking~\cite{fu2022complexity}. Intuitively, LJP appears to be an ideal fit for such models. Legal decision-making often involves comparing multiple candidate charges, evaluating whether each satisfies the legal criteria, and narrowing down to the most appropriate one based on the case facts. As a result, it is natural to expect that strong reasoning models would lead to major improvements in LJP.

However, existing reasoning models fail to deliver the expected breakthroughs in LJP. In practice, they tend to predict the most likely charges without comparing them to similar alternatives, and their reasoning chains are often short and lacking in meaningful intermediate steps. These issues become especially clear in cases involving rare or confusing charges, where accurate judgment depends on subtle distinctions and careful reasoning. Although models may produce step-by-step outputs in such scenarios, the reasoning often stays at a surface level, focusing on pattern matching rather than legal principles.

We argue that the main reason for the limited performance of reasoning models in legal judgment tasks is not a lack of reasoning ability, but a lack of the specialized knowledge that legal reasoning depends on~\cite{yuan2024largelanguagemodelsgrasp}. Effective legal analysis requires long-tail legal knowledge, such as determining the applicability of specific statutes. In some cases, this knowledge is even absent from official legal texts. When such information is missing, models struggle to produce complete and trustworthy reasoning chains as shown in Figure~\ref{fig:intro}. These observations highlight the need for domain-specific knowledge augmentation mechanisms that can dynamically supply essential information during the reasoning process.

To address the knowledge gaps in legal reasoning, we propose GLARE (\textit{A\textbf{G}entic \textbf{L}eg\textbf{A}l \textbf{R}easoning Fram\textbf{E}work}), a modular system that enables language models to dynamically acquire key legal knowledge to improve the breadth and depth of reasoning. First, the \textbf{Charge Expansion Module} (CEM) expands a diverse set of confusing charges by leveraging multiple signals, such as legal structure and historical co-occurrence. This helps the model compare a wider range of candidates and avoid premature conclusions. Second, the \textbf{Precedents Reasoning Demonstration} (PRD) module is built on reasoning paths that are constructed offline from real legal cases. During inference, the model retrieves the most relevant precedents through semantic search and learns from their reasoning chains via in-context learning. Finally, the \textbf{Legal Search-Augmented Reasoning} (LSAR) module allows the model to detect knowledge gaps and retrieve supporting legal information when needed. We guide the model to focus its search on differences between similar charges and details of how specific laws apply, rather than general case facts. Retrieved content is structured and injected into the reasoning process to support more accurate conclusions. By integrating essential legal knowledge, the model achieves more trustworthy and transparent judgment prediction.
 
Following prior work in legal judgment prediction, we conduct experiments on two publicly available real-world legal datasets. Experimental results show that our method consistently outperforms a range of strong baselines. Notably, it achieves substantial improvements on challenging cases involving confusing and difficult charges, where long-tail legal knowledge is crucial. These gains stem from our approach's ability to effectively enrich and incorporate relevant legal knowledge.

In summary, our contributions are as follows:

(1) We introduce GLARE, an agentic framework for legal judgment prediction that enhances reasoning by dynamically integrating legal knowledge throughout the decision-making process.

(2) We design three complementary modules to enrich the model’s reasoning process by expanding candidate charges, leveraging real-world precedents, and injecting retrieved legal knowledge.

(3) Extensive experiments on two real-world datasets show that GLARE significantly outperforms strong baselines, with especially notable gains on cases requiring crucial legal knowledge.

\begin{figure*}[t]
  \includegraphics[width=\linewidth]{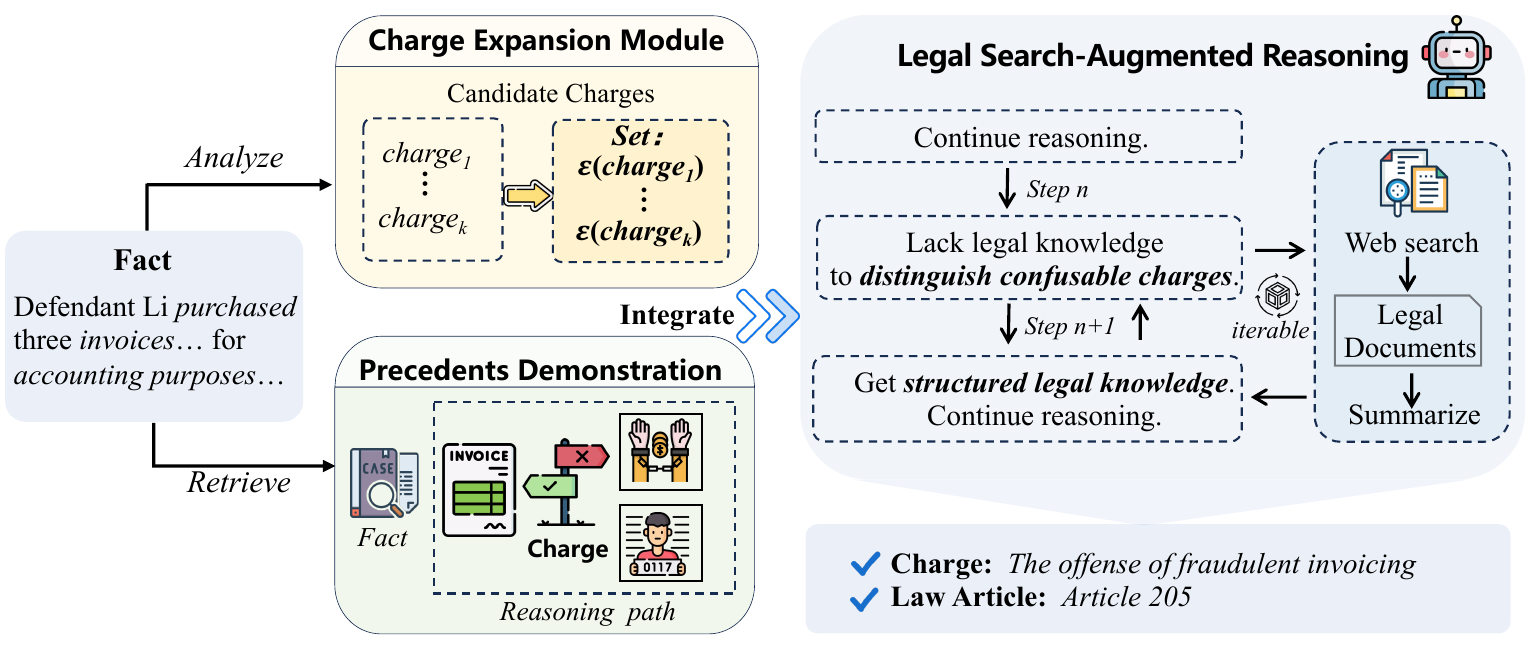}
  \caption{Overview of our agentic legal reasoning framework. LLMs can utilize three external modules to acquire knowledge: Charge Expand Module expands a diverse set of charges, precedents retrieved from offline built database can provide in-context learning, Legal Search-Augmented Reasoning allows the model to detect knowledge gaps and retrieve supporting legal information.}
  \label{fig:overview}
\end{figure*}


\section{Related Work}

\paragraph{Legal judgment prediction}  
Legal judgment prediction has experienced significant development and become an increasingly crucial NLP task. Earlier research~\cite{segal1984predicting} relied on artificially designed features to capture information from legal texts. \citealp{sulea2017exploring} applied traditional machine learning methods to predict the legal judgment. Recent advances in deep learning~\cite{xu2020distinguish,zhang2023case} have motivated researchers to leverage neural networks for automated text representation learning. Recently, LLMs has further promoted the progress of LJP~\cite{deng2024learning}, and several studies~\cite{wu2023precedent,peng2024athena} employ Retrieval-Augmented Generation (RAG) technology~\cite{zhao2024retrieval} to enhance LLMs by incorporating external legal knowledge. However, existing LLM-based methods struggle to utilize comprehensive legal knowledge~\cite{fei2023lawbench} and refer to the way of precedent reasoning to analyze cases. In this context, we make full use of external knowledge and precedents.
\paragraph{Reasoning skills in language models}
Recent work has improved LLMs' reasoning through better prompting techniques~\cite{sahoo2024systematic}. \citet{wei2022chain} showed that chain-of-thought prompting can explicitly guide LLMs to reason step by step. In the legal domain specifically, LoT~\cite{jiang2023legal} proposed legal syllogism reasoning to improve performance on LJP task. ADAPT~\cite{deng2024enabling} further established a comprehensive workflow for LJP that enables discriminative reasoning in LLMs. However, these approaches primarily rely on the LLMs intrinsic capabilities, which inherently constrain the reasoning breadth and the depth of analysis~\cite{zhang2024should,ke2025survey}. Therefore, we propose an agentic legal reasoning framework to dynamically acquire key legal knowledge to improve the breadth and depth of reasoning.

\section{Methodology}

\subsection{Preliminaries}

We first formally define legal judgment prediction. Given a case fact description $f$, the model will analyze and predict the final judgment results including the relevant law articles, the convicted charges and the term of imprisonment for the defendent. Following previous works~\cite{shui2023comprehensive,wei2025llms}, we exclude the task of sentencing prediction from our scope as its subjective nature brings challenges that are not well aligned with the current capabilities of large language models.

In this work, we treat large language models as agentic legal reasoners that can dynamically acquire and incorporate external legal knowledge to enhance their analysis. Rather than relying solely on parametric knowledge, our approach equips the model with access to external modules, enabling it to enrich its reasoning with case-specific legal context. Given a case fact description $f$ and a set of external modules $M$, the model performs step-by-step analysis to construct a coherent reasoning chain $R$ and arrive at a final judgment prediction $p$. We formalize this process as a mapping: $(f, \mathcal{M}) \rightarrow (\mathcal{R}, p)$.

\subsection{Agentic Legal Reasoning Framework}

We propose GLARE, an agentic legal reasoning framework that autonomously invokes external modules to support comprehensive and informed judgment prediction. As shown in Figure~\ref{fig:overview}, GLARE follows a structured three-stage reasoning pipeline:

\begin{enumerate}[leftmargin=10pt]
    \item \textbf{Charge Expansion:} The model begins by analyzing the case facts and generating preliminary candidate charges. To prevent premature narrowing of the decision space, it triggers the Charge Expansion Module to supplement the initial candidates with legally similar charges.

    \item \textbf{Precedent-Enhanced Reasoning:} The model retrieves relevant precedents from an offline-constructed database that includes fact descriptions and synthesized reasoning chains. The reasoning chains were constructed in advance to illustrate the key distinctions between confusing charges. These precedents serve as case-specific reasoning demonstrations, helping the model better understand how similar legal criteria apply and guiding it through more precise reasoning via in-context learning.

    \item \textbf{Iterative Search-augmented Reasoning:} As the model reasons through each candidate charge, it dynamically identifies knowledge gaps such as missing legal definitions and charge-specific thresholds. Rather than treating retrieval as a one-time step, the model interleaves reasoning and retrieval in a loop. Retrieved results are injected back into the reasoning context, enabling the model to refine its current analysis. This iterative process continues until the model has collected sufficient knowledge to complete its reasoning and reach a final judgment.
\end{enumerate}

The three modules collaboratively supplement legal knowledge and extend the legal reasoning chain. Next, we will introduce these three modules in detail.

\subsection{Charge Expansion Module} \label{CEM}

To enable charge comparison and avoid premature conclusions, we expand each candidate charge by retrieving related charges. The expansion is based on two complementary perspectives: legal structure and historical co-occurrence.

\paragraph{Legal Structure-based Expansion.} The Criminal Law is organized into chapters, each representing a specific legal interest or domain. Charges within the same chapter typically differ in subtle legal criteria, while charges across different chapters may involve similar actions or consequences but fall under distinct legal categories. To capture both fine-grained intra-domain distinctions and cross-domain conceptual similarities, we retrieve related charges from both within the same chapter and across different chapters.

Specifically, for a given charge $c$, we use the pretrained dense retriever BGE~\cite{xiao2024c} to find the top-$k$ most similar charges from (a) the same chapter and (b) other chapters:

\begin{equation}
    \begin{aligned}
        \mathcal{E}_1(c) = \text{topk}_{\text{same}}(c) \cup \text{topk}_{\text{diff}}(c),
    \end{aligned}
\end{equation}

where $\text{top-}k_{\text{same}}(c)$ and $\text{top-}k_{\text{diff}}(c)$ represent the most similar charges from the same and different chapters, respectively. This dual-source expansion helps the model compare similar alternatives, reducing the risk of overlooking relevant charges.

\paragraph{History-based Expansion.}

Certain charges tend to appear together in real-world cases, reflecting practical legal dependencies or common joint indictments. We leverage the MultiLJP~\cite{lyu2023multidefendantlegaljudgmentprediction} dataset, where each case may involve multiple defendants and multiple charges. By analyzing these cases, we construct a co-occurrence dictionary that records how frequently each pair of charges appears together. For a given charge $c$, we select the top-$k$ most frequently co-occurring charges as the expansion set $\mathcal{E}_2(c)$.

\paragraph{Final Expansion Set.}
Given an initial set of candidate charges $\{c_1, c_2, \dots\}$ predicted by the language model, we apply the two strategies above to expand each charge:

\begin{equation}
    \begin{aligned}
        \mathcal{E}(c_i) = \mathcal{E}_1(c_i) \cup \mathcal{E}_2(c_i)
    \end{aligned}
\end{equation}

\subsection{Precedents Reasoning Demonstration}

\begin{figure}[t]
  \includegraphics[width=\linewidth]{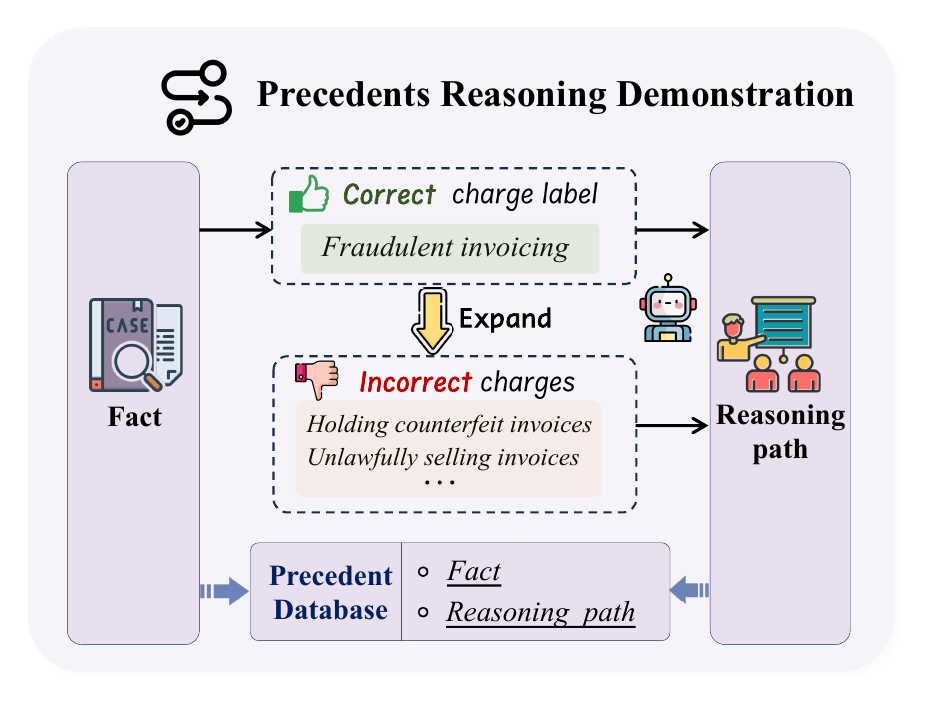}
  \caption{The module of Precedents Reasoning Demonstration: LLM analyzes the reasons for the selection or exclusion of each charge based on facts, thereby generating the reasoning path of precedents.}
  \label{fig:PRD}
\end{figure}

Previous precedent-based approaches~\cite{wu2023precedent,chen2023case,santosh2024incorporating} typically retrieve the fact description and final judgment of prior cases, then insert them directly into the prompt. However, such methods offer little insight into the \textit{reasoning process} behind those decisions. As a result, they tend to rely on shallow fact matching rather than learning how to distinguish between legally similar charges.

To address this issue, we construct reasoning-augmented precedents that make the decision logic explicit. As shown in Figure~\ref{fig:PRD}, we first expand its original charge $c$ into a set of similar charges $C$. 

Given the case fact $f$, the correct charge $c$, and the set of alternatives $C$, we prompt LLM to generate a reasoning path $r$ that explains why $c$ is appropriate and why the other candidates in $C \setminus \{c\}$ should be excluded\footnote{We provide the detailed prompt and examples for synthesizing reasoning paths in Appendix~\ref{PRD prompt}}. This reasoning is generated \textbf{offline} and stored together with the case facts.

\subsection{Legal Search-Augmented Reasoning}

While recent retrieval-augmented generation (RAG) approaches~\cite{wu2023precedent,peng2024athena,feng2024cadlra} enhance legal models by retrieving precedents, statutes, and charge definitions, they remain limited in key aspects. Specifically, they often fail to resolve fine-grained distinctions between similar charges or provide detailed rules to determine facts. Moreover, these methods rely on static retrieval from fixed knowledge bases, making them inflexible and unable to accommodate evolving judicial practices.

To address these limitations, we introduce a \textbf{dynamic and iterative legal search-augmented reasoning mechanism}. Rather than passively injecting generic legal content, our method allows the LLM to actively identify knowledge gaps during reasoning and generate targeted queries. These queries focus on \textit{subtle differences between candidate charges} and \textit{fact-specific questions}. We exclusively source authoritative legal interpretations from official channels, thereby minimizing noise. The system retrieves relevant legal texts from the web in real time, enabling up-to-date and context-related augmentation.

We further ground the model’s reasoning in a \textbf{syllogistic structure}: the retrieved legal context serves as the major premise, the case fact as the minor premise, and the conclusion is derived through logical alignment~\cite{jiang2023legal,HE2025125462}. This structure helps the model remain grounded in factual evidence and reduce hallucinations. The overall reasoning process is formalized as an iterative function:

\begin{equation}
    \begin{aligned}
        R_t = f_\theta(R_{<t}, q_t, d_t, f),
    \end{aligned}
\end{equation}

where $R_t$ denotes the current reasoning state, $R_{<t}$ are the historical reasoning paths, $q_t$ and $d_t$ are the query and corresponding retrieved documents of this step, and $f$ is the case fact.

This design enables the model to incrementally construct a legally grounded reasoning chain, adaptively integrating external knowledge as needed. By decoupling retrieval from static knowledge bases and aligning it with the model's evolving needs, our framework offers greater flexibility to real-world legal dynamics.

\section{Experiments}

\subsection{Datasets and Evaluation}
We conducted experiments in both single-defendant and multi-defendant scenarios to verify the effectiveness of our method in practical applications. For the single-defendant case, we use the CAIL2018 dataset~\cite{xiao2018cail2018largescalelegaldataset}. For the multi-defendant case, we adopt the CMDL dataset~\cite{huang-etal-2024-cmdl}. We uniformly sampled across all charges to construct a balanced test set.
The details are shown in Table~\ref{tab:dataset}. For the PRD module, We employ the training set from both dataset as our precedent database. For evaluation metrics, we adopt the same measures used in prior work: Accuracy (Acc.), \ Macro Precision (Ma-P), \ Macro Recall (Ma-R), \ and Macro F1 (Ma-F).

\begin{table}[h]
  \centering
  \small
  \begin{tabular}{lcc}
  \toprule
  \textbf{Dataset} & \textbf{CAIL2018} & \textbf{CMDL} \\
  \midrule
  \# Train cases & 100,531 & 63,032 \\
  \# Test cases & 1,000 & 834 \\
  \# Charges & 190 & 164 \\
  \# Articles & 175 & 147 \\
  \# Average criminal per case & 1 & 3.79 \\
  Average length per case & 409.6 & 1124.94\\
  \bottomrule
  \end{tabular}
  \caption{Statistics of dataset.}
  \label{tab:dataset}
\end{table}

\begin{table*}[htbp]
  \centering
    \small
  \begin{tabular}{lcccccccc}
    \toprule
    \multirow{2}{*}{Methods} & \multicolumn{4}{c}{Charge} & \multicolumn{4}{c}{Law Article} \\
    \cmidrule(lr){2-5} \cmidrule(lr){6-9} 
    & Acc. & Ma-P & Ma-R & Ma-F & Acc. & Ma-P & Ma-R & Ma-F \\
    \midrule
    \multicolumn{9}{l}{\textit{\textbf{Classification Methods}}} \\
    TopJudge & 52.1 & 50.9 & 45.7 & 43.5 & 52.8 & 47.7 & 43.8 & 41.2 \\
    LADAN & 76.7 & 73.4 & 71.0 & 69.5 & 77.5 & 71.0 & 69.2 & 67.5 \\
    NeurJudge & 74.7 & 77.7 & 71.5 & 71.5 & 77.4 & 80.7 & 74.6 & 74.3 \\
    BERT & 85.8 & 83.4 & 86.6 & 83.3 & 85.8 & 80.4 & 82.8 & 79.9 \\
    Lawformer & 71.3 & 58.2 & 62.7 & 57.8 & 72.9 & 58.1 & 61.4 & 56.9 \\
    \midrule
    \multicolumn{9}{l}{\textit{\textbf{Direct Reasoning}}} \\
    Qwen2.5-32B & 74.5 & 75.3 & 69.3 & 69.1 & 77.1 & 73.3 & 66.6 & 67.1  \\
    QwQ-32B & 82.5 & 86.9 & 80.5 & 80.9 & 84.0 & 83.1 & 76.1 & 77.0 \\
    Qwen2.5-72B & 76.6 & 78.9 & 72.2 & 72.3 & 77.7 & 73.4 & 66.8 & 67.3 \\
    DeepSeek-R1-671B & 84.8 & 86.3 & 81.3 & 81.7 & 87.2 & 86.8 & 81.8 & 82.6 \\
    \midrule
    \multicolumn{9}{l}{\textit{\textbf{Retrieval-augmented Reasoning}}} \\
    
    Precedent-based-RAG-Qwen2.5-32B & 88.5 & 88.2 & 85.8 & 85.7 & 89.4 & 87.2 & 83.7 & 84.5 \\
    Precedent-based-RAG-QwQ-32B & 89.4 & 89.9 & 87.3 & 87.1 & 90.4 & 88.4 & 85.2 & 85.4  \\
    Precedent-based-RAG-Qwen2.5-72B & 88.1 & 87.5 & 85.1 & 84.9 & 89.4 & 86.8 & 83.9 & 84.0 \\
    Search-o1-QwQ-32B & 81.8 & 85.3 & 78.8 & 79.3 & 83.9 & 83.3 & 76.4 & 77.4 \\
    \midrule
    \multicolumn{9}{l}{\textit{\textbf{Agentic Retrieval-augmented Reasoning}}} \\
    \rowcolor[RGB]{236,244,252}
    GLARE-Qwen2.5-32B(ours) & \textbf{89.8} & 89.8 & 87.8 & 87.8 & 90.4 & 89.2 & 87.3 & 87.5 \\ 
    \rowcolor[RGB]{236,244,252} 
    GLARE-QwQ-32B(ours) & 89.7 & \textbf{90.7} & \textbf{88.6} & \textbf{88.6} & \textbf{91.3} & \textbf{90.6} & \textbf{88.3} & \textbf{88.5} \\
    \bottomrule
  \end{tabular}
  \caption{Performance comparison on CAIL2018 dataset. The best results are in bold.}
  \label{tab:cail2018 performance}
\end{table*}
\subsection{Baselines}
We compare our method against two categories of baseline approaches:

\paragraph{Classification Methods:}
These methods take legal judgment prediction as a classification task, relying on supervised learning with labeled datasets. \textbf{TopJudge}~\cite{zhong2018legal} employs a graph structure to model the topological dependency among the three subtasks: charge prediction, law article prediction, and sentence term prediction. \textbf{NeurJudge}~\cite{yue2021neurjudge} integrates a legal knowledge graph into the neural architecture, capturing explicit relationships among legal entities and improving reasoning over structured legal knowledge. \textbf{BERT}~\cite{devlin2019bert}, a standard pre-trained transformer model, is adapted to legal texts via supervised training. It serves as a strong baseline for judgment prediction tasks. \textbf{Lawformer}~\cite{xiao2021lawformer} is built upon Longformer~\cite{beltagy2020longformer} and further pretrained on large-scale Chinese legal corpora, which enhances its ability to process longer legal documents and capture complex contextual semantics.  

\paragraph{LLM-based Methods:}
These methods utilize LLMs to perform legal reasoning in zero-shot or few-shot settings~\cite{brown2020language}. \textbf{Direct Reasoning} directly feeds the case facts into the LLM to predict the applicable law articles and charges, without relying on any retrieval augmentation or additional external context. The models used in this setting include Qwen2.5-32B/72B-Instruct~\cite{yang2024qwen2}, QwQ-32B~\cite{qwq32b}, and DeepSeek-R1-671B~\cite{guo2025deepseek}. \textbf{Retrieval-augmented Reasoning:} \textit{(1) Precedent-based RAG} enhances reasoning by retrieving top-5 precedents including their facts and labels, which are appended to the prompt. The models used in this setting include Qwen2.5-32B/72B-Instruct~\cite{yang2024qwen2}, QwQ-32B~\cite{qwq32b}. \textit{(2) Search-o1}~\cite{li2025search} dynamically retrieves external knowledge when it encounters uncertain or ambiguous knowledge in the general domain. We use reasoning model QwQ-32B~\cite{qwq32b} in this setting.

\begin{table*}[h]
  \centering
  \small
  \begin{tabular}{lcccccccc}
    \toprule
    \multirow{3}{*}{Methods} 
    & \multicolumn{4}{c}{CAIL2018} & \multicolumn{4}{c}{CMDL} \\
    \cmidrule(lr){2-5}\cmidrule(lr){6-9}
    & \multicolumn{2}{c}{Charge} & \multicolumn{2}{c}{Law Article} & \multicolumn{2}{c}{Charge} & \multicolumn{2}{c}{Law Article} \\
    \cmidrule(lr){2-3} \cmidrule(lr){4-5} \cmidrule(lr){6-7} \cmidrule(lr){8-9}
    & Acc. & Ma-F & Acc. & Ma-F & Acc. & Ma-F & Acc. & Ma-F \\
    \midrule
    \multicolumn{9}{l}{\textit{\textbf{Direct Reasoning}}} \\
    Qwen2.5-32B & 60.2 & 39.3 & 63.7 & 41.2 & 57.4 & 64.7 & 57.9 & 63.7 \\
    QwQ-32B & 78.4 & 57.0 & 79.2 & 58.2 & 67.9 & 72.9 & 69.8 & 74.2 \\
    \midrule
    \multicolumn{9}{l}{\textit{\textbf{Retrieval-augmented Reasoning}}} \\
    Precedent-based-RAG-Qwen2.5-32B & 82.6 & 62.7 & 83.0 & 62.3 & 65.7 & 69.5 & 65.3 & 67.6  \\
    Precedent-based-RAG-QwQ-32B & 84.6 & 65.5 & 84.6 & 67.6 & 72.8 & 74.8 & 71.3 & 73.2  \\
    \midrule
    \multicolumn{9}{l}{\textit{\textbf{Agentic Retrieval-augmented Reasoning}}} \\
    \rowcolor[RGB]{236,244,252}
    GLARE-Qwen2.5-32B(ours) & 86.9 & 68.6 & 86.5 & 68.3 & 73.5 & 75.5 & 71.9 & 73.4 \\ 
    \rowcolor[RGB]{236,244,252}
    GLARE-QwQ-32B(ours) & \textbf{90.7} & \textbf{75.7} & \textbf{91.1} & \textbf{75.4} & \textbf{76.0} & \textbf{79.5} & \textbf{74.0} & \textbf{76.7} \\
    \bottomrule
  \end{tabular}
  \caption{Performance comparison on difficult charges.}
  \label{tab:difficult charges}
\end{table*}

\subsection{Experiment Settings}

In our experiments, we adopt Qwen2.5-32B~\cite{yang2024qwen2} and QwQ-32B~\cite{qwq32b} as the base models to run the full reasoning pipeline. For generation, we set the following parameters: a maximum of 32,768 tokens and temperature of 0.6. For charge expansion, we set the top-\textit{k} expanded charges to 3 in each expansion method. For precedent retrieval, we use SAILER~\cite{li2023sailer} to encode case facts and set the top-\textit{k} retrieved precedents to 5. In the legal search module, we utilize Serper API~\footnote{\url{https://serper.dev}} with the region configured for China and the number of returned results limited to the top 10. For charges that are not in the predefined label set, we map them to the most similar charge within the label set using BGE~\cite{xiao2024c}.
\subsection{Experiment Results}
    
The results are reported in Table~\ref{tab:cail2018 performance} and Appendix~\ref{CMDL}, and next we will analyze the experimental results:

\textbf{1. Our method has demonstrated consistent performance improvements in both charge prediction and law article prediction tasks,} highlighting the effectiveness of our agentic reasoning approach for LJP. Compared to the Direct Reasoning setting, our method improves charge prediction by 7.7\% and law article prediction by 11.5\% in F1 score. When compared with Retrieval-augmented Reasoning, it achieves an improvement of 1.5\% on charge prediction and 3.1\% on law article prediction in F1 score. In addition, our method not only performs well on the large reasoning models, but also effectively promotes the reasoning ability of the instruct models, indicating that our three modules effectively supplement legal knowledge and thereby enhance reasoning performance.

\textbf{2. In contrast to direct reasoning, precedent-based RAG enhances prediction performance through precedent retrieval.} Large reasoning models like QwQ-32B and DeepSeek-R1-671B outperform other instruct models in direct reasoning, indicating that LJP inherently requires multi-step reasoning and slow thinking. Precedent-based RAG improves performance across models of various scales by incorporating precedent retrieval. For example, QwQ-32B sees an 8.39\% F1 improvement in law article prediction. However, precedent-based RAG only provides the case facts and labels of precedents, leading models to rely on similarity matching and copy judgment predictions rather than truly reason. Additionally, Search-o1 retrieves case facts which may introduce noises, rather than specific legal knowledge, thus underperforming compared to direct reasoning.

\textbf{3. LLM-based methods outperform classification methods.} However, BERT demonstrates superior performance compared to LLMs such as Qwen2.5-72B-Instruct in the direct reasoning setting, although the latter have significantly larger parameter sizes. The key reasons are as follows: (1) BERT frames charge and article prediction as multi-class classification tasks, enabling direct mapping from facts to fixed labels, which aligns well with the task. In contrast, LLMs take a generative approach and without legal-specific training they often fail to make accurate predictions. (2) Our dataset includes many rare and confusing charges. Fine-tuned BERT models trained on legal corpora can better distinguish these nuanced charges, while even large LLMs lack the domain knowledge needed for such difficult charges.

\subsection{Ablation Study}

\begin{table}[htbp]
  \centering
  \small
  \begin{tabular}{lcccccccc}
    \toprule
    \multirow{2}{*}{Methods} & \multicolumn{2}{c}{Charge} & \multicolumn{2}{c}{Law Article} \\
    \cmidrule(lr){2-3} \cmidrule(lr){4-5} 
    & Acc. & Ma-P & Acc. & Ma-F \\
    \midrule
    w/o CEM & 89.6 & 87.7 & 90.3 & 85.2 \\
    w/o PRD & 80.0 & 78.1 & 81.6 & 75.4 \\
    w/o LSAR & 89.6 & 87.9 & 90.4 & 86.5 \\
    GLARE(ours) & \textbf{89.7} & \textbf{88.6} & \textbf{91.3} & \textbf{88.5} \\
    \bottomrule
  \end{tabular}
  \caption{Ablation Study. The best results are in bold.}
  \label{tab:ablation}
\end{table}

To evaluate the effectiveness of each component in the GLARE framework, we conducted ablation experiments with the following strategies: \textbf{(1) w/o CEM:} The Charge Expansion Module is removed, so the model cannot expand a diverse set of candidate charges. \textbf{(2) w/o PRD:} The Precedents Reasoning Demonstration module is removed, so the model cannot refer to reasoning path from precedents. \textbf{(3) w/o LSAR:} The Legal Search-Augmented Reasoning module is removed, disabling the model’s ability to supplement its knowledge via external legal search when faced with ambiguous or unfamiliar charges.

\begin{figure*}[t]
  \includegraphics[width=\linewidth]{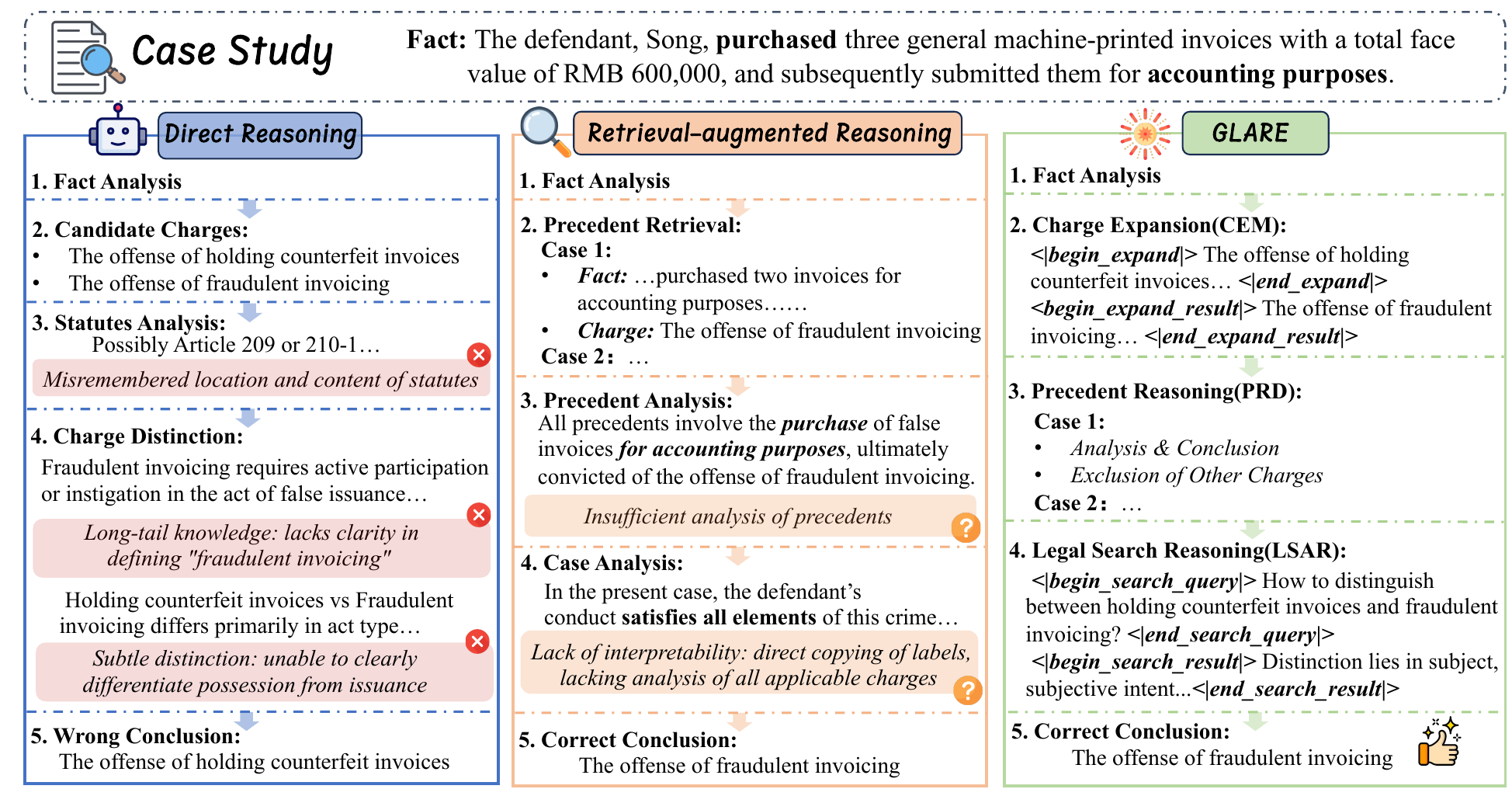}
  \caption{Case Study. The red part highlights the model's limitations due to insufficient internal knowledge, while the yellow part demonstrates the lack of interpretability in vanilla precedent-based RAG reasoning.}
  \label{fig:case}
\end{figure*}

As shown in Table~\ref{tab:ablation}, the removal of any single module results in degraded performance. In particular, removing PRD causes the most significant degradation: the accuracy of charge prediction drops from 89.7\% to 80\%. This highlights the crucial role of precedent reasoning path in enhancing legal judgment prediction. Removing CEM weakens the model's ability to recognize ambiguous or low-frequency charges, while LSAR helps the model fill knowledge gaps by retrieving authoritative legal information. Overall, the GLARE framework performs best across all metrics, validating the strength of agentic reasoning in legal judgment prediction.

\subsection{Efficiency Analysis} 

\begin{figure}[t]
  \includegraphics[width=\linewidth]{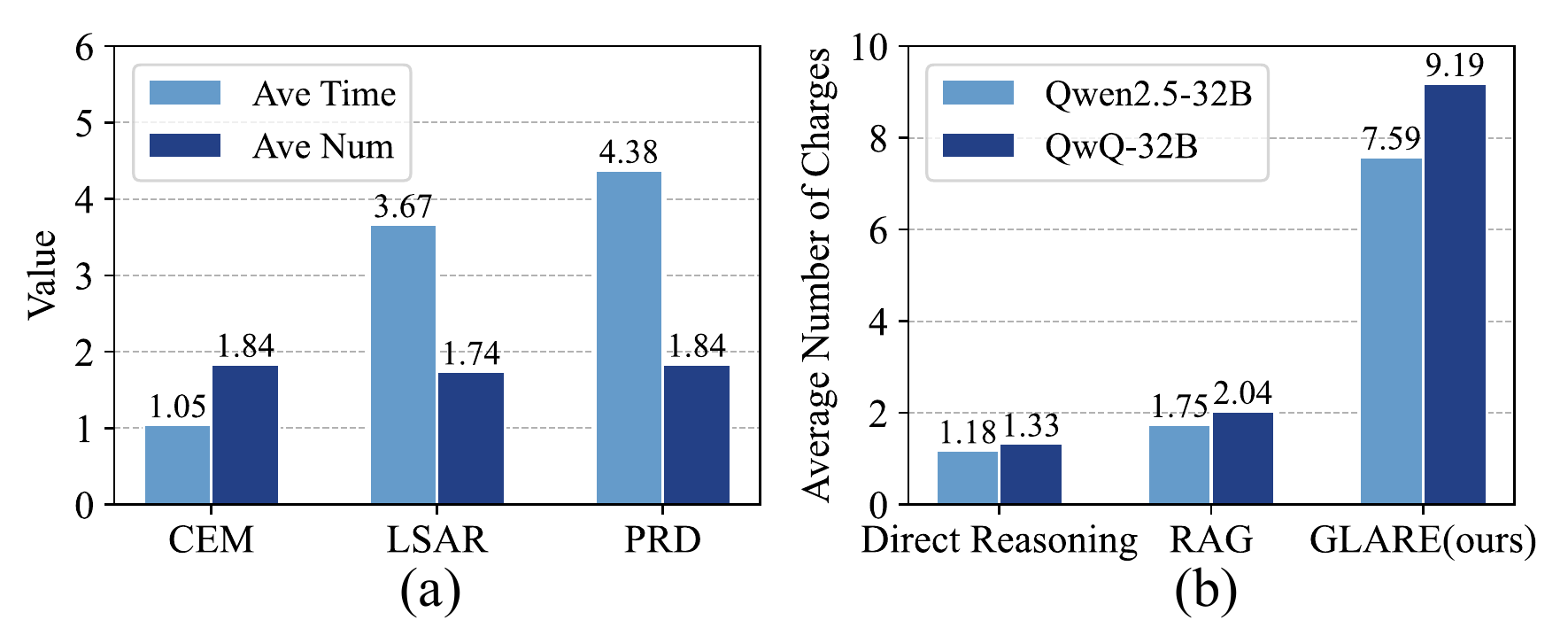}
  \caption{(a) Efficiency analysis of each module. (b) Average charge numbers of different methods.}
  \label{fig:charge num}
\end{figure}

In this work, we focus on multi-step reasoning and slow thinking for legal judgment prediction, so the latency is less important. Nevertheless, we still conducted an analysis to further understand each module. Based on the analysis of Figure~\ref{fig:charge num} (a), we can draw the following conclusions:

\textbf{(1) The overall inference efficiency is relatively high.} The average reasoning rounds for each case is 5.17 and the average call numbers for each module is between 1.7 and 1.8 times, indicating that the module scheduling is well-balanced, without obvious redundancy or repeated invocation. So the overall delay is within our acceptable range.

\textbf{(2) The CEM module is the most efficient.} In both expansion methods, the charge structures are established offline in advance, so its computation cost is low and runtime is minimal. As shown in the figure~\ref{fig:charge num} (b), compared with direct reasoning and RAG approaches, our method considers a comprehensive set of charges and performs a more thorough analysis. 

\textbf{(3) The PRD module has the highest latency but within an acceptable range.} Since this module needs to encode the entire case description and the case text is usually long, the reasoning time is relatively long. However, the PPR module can provide the reasoning path of precedents and has significant reasoning interpretability.

\subsection{Case Study}

As shown in the Figure~\ref{fig:case}, we conducted case study on three LLM-based methods to further verify the effectiveness and interpretability of our method. Direct Reasoning relies on the LLM’s internal knowledge, which may be inaccurate or insufficient, leading to incorrect judgments. RAG methods often lack explicit links between retrieved cases and final decisions, making it hard to trace how external knowledge affects reasoning. However, our method ensures that each reasoning step has a clear knowledge basis through the explicit invocation of three modules, thereby extending the reasoning chain.

\subsection{Performance on Difficult Charges}
To evaluate GLARE's ability to handle challenging charges requiring long-tail knowledge, We conducted experiments on low-frequency charges with less than 100 cases (e.g., the crime of bribing a unit) and confusing charges (e.g., robbery vs. snatching). The results are reported in Table~\ref{tab:difficult charges}, which reveal two key insights: (1) Our method dynamically acquires critical legal knowledge, outperforming Direct Reasoning by over 10\% and Retrieval-augmented Reasoning by over 5\%. (2) RAG-based methods struggle to retrieve relevant precedents for such charges, leading to poor performance, while direct reasoning fall short due to limited long-tail knowledge. These results highlight the strength of our external modules in supplementing legal reasoning with critical knowledge.

\section{Conclusion}
In this study, we propose a novel framework, GLARE, to address the legal gaps in legal reasoning. GLARE dynamically acquires key legal knowledge to improve the breadth and depth of reasoning. Experimental results demonstrate the effectiveness of our approach, which not only improves prediction performance but also generates complete reasoning chains that enhance the interpretability of LJP tasks. We believe that GLARE holds great potential for real-world legal applications and will contribute meaningfully to the advancement of intelligent judicial systems.

\section*{Limitations}

\paragraph{Generalizability}
We adopted the legal dataset from China Judgments Online to verify the applicability of the method in the China's judicial system. However, the GLARE framework is applicable to countries following both common law and civil law systems. When applied to the actual judicial practice of a specific country, we need to inject the specific legal knowledge base of each country and adapt to the local judicial culture.

\paragraph{Efficiency}
Our method promotes the reasoning ability of the model through multiple rounds of reasoning and the invocation of three modules. Although this process has an increased time cost compared to the traditional direct reasoning method, the task of legal judgment prediction itself is a task that requires multi-step reasoning and slow thinking. Moreover, this time cost is much less than the time needed for humans to analyze cases in real life. Therefore, such a time cost is acceptable.

\section*{Ethical Discussion}
\paragraph{Potential Bias in Legal Data}
Large language models may learn historical bias from legal judgments in training data. In practice, judicial decisions may be influenced by many external factors, such as public opinion, regional differences or the personal inclinations of judges. We need to identify possible biases before deploying such models in real-world scenarios.

\paragraph{Human-Centric Deployment}
Our system is designed to assist judges by providing supplementary recommendations rather than replacing human decision-making. We advise users to critically evaluate the model’s predictions and make independent decisions about their adoption, rather than uncritically accepting the model’s reasoning.

\bibliography{main}

\clearpage
\appendix
\section*{Appendix}

\section{Instructions of Invoking External Modules}

(a) Charge Expansion Module and Precedents Reasoning Demonstration: We instruct LLMs to generate preliminary candidate charges enclosed within special symbols \green{<|begin\_expand|>} and \green{<|end\_expand|>}. When such symbols are detected, LLMs will stop reasoning and invoke Charge Expansion Module to expand charges. The expanded charges are injected back into the reasoning chain along with retrieved precedents. (b) Legal Search-Augmented Reasoning: Similarly, we instruct LLMs to generate queries encapsulated between special symbols \green{<|begin\_search|>} and \green{<|end\_search|>} to trigger the retrieval mechanism. The system then invokes a web search to obtain relevant information. Detailed instructions are shown in Figure~\ref{fig:instructions}.

\section{Prompts of GLARE}

The main analysis process is: (1) LLMs initially analyze the case and generate preliminary candidate charges, then expand a diverse set of similar charges; (2) retrieve the most relevant precedents; (3) integrate the retrieved precedents and the expanded charges into the reasoning chain together, and then analyze each candidate charge via in-context learning; (4) during the reasoning process, when encountering knowledge gaps, autonomously acquire structured legal knowledge through web search, ultimately outputting applicable law articles and convicted charges. Detailed prompts are shown in Figure~\ref{fig:glare-prompt}.

\section{Prompts of Precedents Reasoning Demonstration} \label{PRD prompt}

We construct reasoning-augmented precedents that make the decision logic explicit. We first expand the original charge label into a set of similar charges. Then given the case fact $f$, the correct charge $c$, and the set of alternatives $C$, we prompt LLM to generate a reasoning path $r$. Detailed prompts are shown in Figure~\ref{fig:The prompt used for PRD}. Next, we present a specific example of precedent reasoning path:

\textit{\textbf{Case Description:} From Nov 2011 to May 2012, defendant Song purchased 2 forged tax invoices (total ¥441,563) to document rent payments from Qingdao Gaoshi Thread Co., Ltd. Forensic examination confirmed all invoices were counterfeit.}

\textit{\textbf{Analysis and Conclusion:} The correct Charge: Issuing False Invoices (Article 205-1 of Criminal Law). The defendant's conduct satisfies the essential elements of issuing false invoices: According to Article 205-1, this crime involves illegally issuing ordinary invoices for others, for oneself, having others issue invoices for oneself, or introducing others to issue invoices. Song purchased forged invoices and provided them to the company for accounting purposes, constituting typical "issuing false invoices" behavior. The intent was not merely possessing counterfeit invoices but fabricating transaction records for financial fraud.} 

\textit{\textbf{Exclusion of Other Charges:} \textbf{(1) Possession of Forged Invoices:} While Song did possess forged invoices, this was merely the means to commit invoice fraud. According to the Doctrine of Implicated Offense in criminal theory: When one act violates multiple charges, the more severe offense applies. Issuing false invoices carries heavier penalties (maximum 7-year imprisonment) than possession (maximum 2 years). \textbf{(2) Issuing Special VAT Invoices:} Only applies to VAT invoices or other tax-deductible invoices (Article 205). The case involves ordinary invoices without tax-related functions. \textbf{(3) Financial Instrument Fraud:} Requires intent to illegally possess others' property through forged instruments (Article 194). Song aimed to document rent payments, not directly defraud. \textbf{(4) False Registered Capital Reporting:} Concerns fraudulent capital contributions during company establishment (Article 158). Irrelevant to invoice-related conduct. \textbf{(5) Illegal Sale of Invoices:} Applies to selling genuine invoices (Article 207). Song purchased rather than sold invoices. }

\textit{\textbf{Conclusion:} Defendant Song's conduct constitutes \textbf{Issuing False Invoices} (Article 205-1) as it fulfilled all statutory elements with greater social harm than alternative charges. Other charges were excluded due to: (1) mismatched conduct objects, (2) different subjective intents, or (3) being secondary implicated offenses.}

\section{Results of CMDL dataset} \label{CMDL}
The results are reported in Table~\ref{tab:cmdl2018 performance}. Our method demonstrated the best performance on both tasks and even outperformed the powerful DeepSeek-R1. Compared with direct reasoning and static RAG, GLARE can autonomously and dynamically supplement knowledge, thereby extending the reasoning chain and enhancing interpretability.
\begin{table*}[htbp]
  \centering
  \begin{tabular}{lcccccccc}
    \toprule
    \multirow{2}{*}{Methods} & \multicolumn{4}{c}{Charge} & \multicolumn{4}{c}{Law Article} \\
    \cmidrule(lr){2-5} \cmidrule(lr){6-9} 
    & Acc. & Ma-P & Ma-R & Ma-F & Acc. & Ma-P & Ma-R & Ma-F \\
    \midrule
    \multicolumn{9}{l}{\textit{\textbf{Direct Reasoning}}} \\
    Qwen2.5-32B & 70.4 & 76.0 & 80.2 & 77.2 & 73.5 & 79.3 & 81.0 & 79.3  \\
    QwQ-32B & 77.9 & 81.7 & 83.7 & 82.1 & 79.5 & 83.4 & 85.1 & 83.5  \\
    Qwen2.5-72B & 73.6 & 77.5 & 80.6 & 78.5 & 76.1 & 80.1 & 80.9 & 79.8  \\
    DeepSeek-R1-671B & 81.4 & 84.1 & 84.9 & 84.1 & 83.3 & 87.4 & 88.4 & 87.2  \\
    \midrule
    \multicolumn{9}{l}{\textit{\textbf{Retrieval-augmented Reasoning}}} \\
    
    Precedent-based-RAG-Qwen2.5-32B & 82.9 & 85.9 & 87.2 & 86.2 & 82.1 & 85.7 & 86.2 & 85.4  \\
    Precedent-based-RAG-QwQ-32B & 83.6 & 86.1 & 87.2 & 86.3 & 82.3 & 85.6 & 86.0 & 85.3  \\
    Precedent-based-RAG-Qwen2.5-72B & 83.2 & 86.3 & 88.1 & 86.7 & 82.2 & 85.9 & 87.1 & 85.9  \\
    \midrule
    \multicolumn{9}{l}{\textit{\textbf{Agentic Retrieval-augmented Reasoning}}} \\
    \rowcolor[RGB]{236,244,252}
    GLARE-Qwen2.5-32B(ours) & 85.4 & 88.1 & 89.1 & 88.2 & 84.5 & 88.1 & 87.8 & 87.4  \\ 
    \rowcolor[RGB]{236,244,252} 
    GLARE-QwQ-32B(ours) & \textbf{86.5} & \textbf{88.8} & \textbf{89.5} & \textbf{88.8} & \textbf{86.2} & \textbf{89.0} & \textbf{89.6} & \textbf{88.8}  \\
    \bottomrule
  \end{tabular}
  \caption{Performance comparison on CMDL dataset. The best results are in bold.}
  \label{tab:cmdl2018 performance}
\end{table*}

\begin{figure*}[t]
\centering
\begin{tcolorbox}[title=User prompts for GLARE, colback=gray!5!white, colframe=gray!75!black, fonttitle=\bfseries, width=\textwidth]
Your task involves predicting the applicable law articles and charges for a given case description. Based on the case facts, you should identify the most relevant law article and charge using your legal knowledge and external knowledge.

\textbf{Procedure:}
\begin{itemize}
    \item Carefully analyze the provided case description.
    \item Generate all possible preliminary predictions.
    \item Use available tools to expand the list of candidate charges, obtaining a more comprehensive set of potential charges along with reasoning paths from precedents.
    \item Analyze each candidate charge's applicability to the current case using syllogistic reasoning, referencing the reasoning paths from similar cases.
    \item When multiple charges remain candidates, conduct web searches to clarify distinctions between them (e.g., ``How to distinguish between Negligent Homicide and Gross Responsibility Accident Crime?'').
    \item Continue the syllogistic analysis by combining search results with case facts. If search results contain unclear information, perform additional searches as needed (e.g., ``What constitutes production operations in Gross Responsibility Accident Crime?''), ultimately determining the most appropriate legal provision and charge.
\end{itemize}

\textbf{Case description:} \texttt{\{fact\}}\\
\textbf{Defendant:} \texttt{\{criminal\}}

\textbf{Important Notes:}\\
All reasoning should be strictly grounded in the facts of the present case.
\end{tcolorbox}
\caption{The prompt used for GLARE.}
\label{fig:glare-prompt}
\end{figure*}

\begin{figure*}[t]
\centering
\begin{tcolorbox}[title=Instructions of Invoking External Modules, colback=gray!5!white, colframe=gray!75!black, fonttitle=\bfseries, width=\textwidth]
You are a reasoning assistant with the ability to call external tools. You have access to the following special tools:\\

\textbf{Expansion Tool:} Expands the initially predicted candidate charges. Write \texttt{<|begin\_expand|> Charge 1, Charge 2, ... <|end\_expand|>}.\\
The system will use existing tools to expand the charges you provide, obtaining a more comprehensive list of candidate charges while also supplying the reasoning paths from similar cases. The results will be returned in the format:\\
\texttt{<|begin\_expand\_result|> Expanded charges, Reasoning paths from precedents <|end\_expand\_result|>}.\\
You may use the expansion tool only once, and it must be called immediately after obtaining the initial predicted candidate charges.\\

\textbf{Search Tool:} When encountering multiple matching charges, performs a web search to clarify the distinctions between them. Write \texttt{<|begin\_search\_query|> How to distinguish between [Charge A] and [Charge B] <|end\_search\_query|>}.\\
Next, the system performs a search and analysis of web pages, outputting the information in the specified format:\\
\texttt{<|begin\_search\_result|> ...Search results... <|end\_search\_result|>}.\\
You may perform multiple searches, with a maximum attempt limit of \texttt{\{MAX\_SEARCH\_LIMIT\}}.\\

\textbf{Example:}\\
Assistant:\\
I need to expand the charges of ``Causing Death by Negligence,'' ``Endangering Public Security by Negligent Use of Dangerous Methods,'' and ``Gross Responsibility Accident Crime.''\\
Assistant:\\
\texttt{<|begin\_expand|> Causing Death by Negligence, Endangering Public Security by Negligent Use of Dangerous Methods, Gross Responsibility Accident Crime <|end\_expand|>}\\
(System returns expansion results)\\
\texttt{<|begin\_expand\_result|> Expanded charges, Reasoning paths from similar cases <|end\_expand\_result|>}\\

Assistant:\\
I need to identify the key differences between ``Causing Death by Negligence'' and ``Gross Responsibility Accident Crime.''\\
Assistant:\\
\texttt{<|begin\_search\_query|> Distinguishing between Causing Death by Negligence and Gross Responsibility Accident Crime <|end\_search\_query|>}\\
(System returns search results)\\
\texttt{<|begin\_search\_result|> ...Search results... <|end\_search\_result|>}
\end{tcolorbox}
\caption{The instruction used for GLARE.}
\label{fig:instructions}
\end{figure*}

\begin{figure*}[t]
\centering
\begin{tcolorbox}[title=Prompts of Precedents Reasoning Demonstration, colback=gray!5!white, colframe=gray!75!black, fonttitle=\bfseries, width=\textwidth]
Given a description of a legal case along with the correct charge, as well as other potential charges, analyze the reasons for selecting the correct charge while excluding the others.\\
1. Case description: \texttt{\{fact\}} \\
2. Correct charge: \texttt{\{charge\_label\}} \\
3. Other potential charges: \texttt{\{expanded\_crimes\}}
\end{tcolorbox}
\caption{The prompt used for PRD.}
\label{fig:The prompt used for PRD}
\end{figure*}

\end{document}